\RequirePackage{fix-cm}
\documentclass{svjour3}
\smartqed
    
\usepackage{algorithm}
\usepackage{algorithmic}
\usepackage{amssymb,amsmath}
\usepackage{enumerate}
\usepackage{float}
\usepackage{ifpdf}
\usepackage{multirow}
\usepackage{subfigure}
\usepackage{verbatim}
\usepackage{url}
\usepackage{epstopdf}
\usepackage{epsfig}
\usepackage[german,english]{babel}
\usepackage{float}
\usepackage{rotating}

\newcommand{\nms}{\negmedspace}
\newcommand{\csac}{{Score \& Combine}}
\newcommand{\ccas}{{Combine \& Score}}
\newcommand{\sac}{{score \& combine}}
\newcommand{\cas}{{combine \& score}}
\journalname{Machine Learning}

\begin{document}

\title{Evaluating Classifiers Without Expert Labels}

\author{
Hyun Joon Jung \and
Matthew Lease
}

\institute{Hyun Joon Jung \at
	School of Information\\
	University of Texas at Austin\\
	\email{hyunJoon@utexas.edu}        
     	\and
	Matthew Lease \at
	School of Information\\
	University of Texas at Austin\\
        \email{ml@ischool.utexas.edu}
}
%\date{Received: August, 10, 2012 / Accepted: ~}

\maketitle
\begin{abstract}
This paper considers the challenge of evaluating a set of classifiers, as done in shared task evaluations like the KDD Cup or NIST TREC, without expert labels. While expert labels provide the traditional cornerstone for evaluating statistical learners, limited or expensive access to experts represents a practical bottleneck. Instead, we seek methodology for estimating performance of the classifiers (relative and absolute) which is more scalable than expert labeling yet preserves high correlation with evaluation based on expert labels. We consider both: 1) using only labels automatically generated by the classifiers themselves ({\em blind evaluation}); and 2) using labels obtained via crowdsourcing. While crowdsourcing methods are lauded for scalability, using such data for evaluation raises serious concerns given the prevalence of label noise. In regard to blind evaluation, two broad strategies are investigated: \emph{\cas} and \emph{\sac}. \emph{\ccas} methods infer a single ``pseudo-gold'' label set by aggregating classifier labels; classifiers are then evaluated based on this single pseudo-gold label set. On the other hand, \emph{\sac} methods: i) sample multiple label sets from classifier outputs, ii) evaluate classifiers on each label set, and iii) average classifier performance across label sets. When additional crowd labels are also collected, we investigate two alternative avenues for exploiting them: 1) direct evaluation of classifiers; or 2) supervision of \emph{combine-and-score} methods. To assess generality of our techniques, classifier performance is measured using four common classification metrics, with statistical significance tests establishing relative performance of the classifiers for each metric. Finally, we measure both score and rank correlations between estimated classifier performance vs.\ actual performance according to expert judgments. Rigorous evaluation of classifiers from the TREC 2011 Crowdsourcing Track shows reliable evaluation can be achieved without reliance on expert labels.

\end{abstract}

%==============================================================================
\section{Introduction} \label{sec:introduction}

\begin{figure*} 
\centering
\ifpdf %ML
\includegraphics[height=50mm]{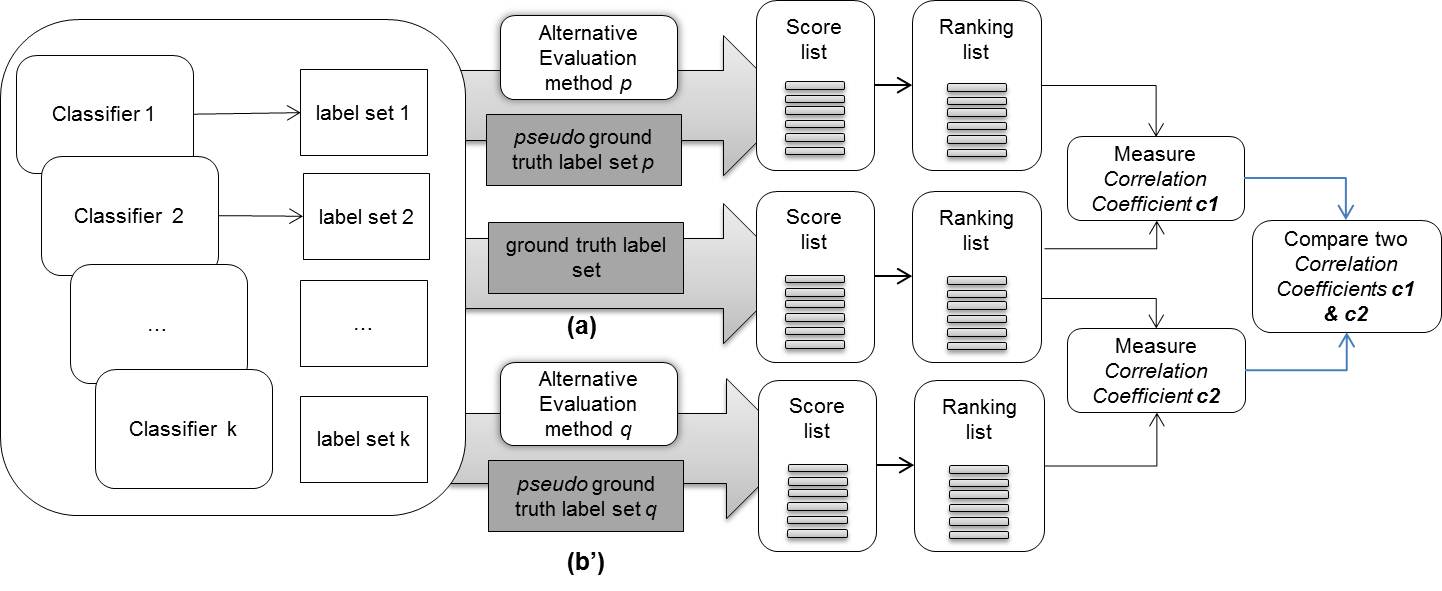}
\fi
\caption{Our experimental framework used. As input, $K$ binary classifiers each label $M$ examples. (a) As ground truth, classifiers are scored for several metrics based on expert judgments, statistical significance of differences is computed, and classifiers are ranked (best to worst). (b) An estimation method {\em p} is used to predict classifier scores without expert judgments, and classifiers are ranked accordingly to estimated scores (score differences which are not statistically significant yield tied rankings). Score and rank correlation is then measured between estimated vs.\ actual scores and ranks ({\em c1}). (b') A second, alternative method {\em q} is used to estimate classifier scores, classifiers are ranked accordingly, and correlation of scores and ranks vs.\ ground truth is measured ({\em c2}). Finally, we compare the correlations {\em c1} and {\em c2} to determine whether {\em p} or {\em q} achieved the greatest score and rank correlation (with statistical significance).}
\label{fig:overall}
\end{figure*}

While expert labels provide the traditional cornerstone for evaluating statistical learners, limited or expensive access to experts represents a practical bottleneck. To be more specific, consider the field of Information Retrieval (IR), where expert labels in the form of {\em relevance judgments} provide the foundation for evaluating IR systems under the Cranfield paradigm~\cite{Cleverdon66}. Such labels enable evaluation of IR systems for both ranking and classification scenarios: ranking items in a standing collection for {\em ad hoc} queries, or classifying new items as they arrive based on standing queries (e.g., RSS filtering). In order to accurately characterize system effectiveness, experience has shown that IR systems should be evaluated at the operational scale at which they will be used in practice. However, as collections sizes have rapidly grown in recent years, it has become increasingly infeasible to manually label so many examples using traditional expert labeling. In at least one case, insufficient labels have compromised a NIST TREC shared task evaluation~\cite{Voorhees06}. As such, the IR community has become particularly interested in developing more scalable evaluation methodology. While statistical sampling techniques and robust evaluation metrics have significantly reduced the number of examples needing to be labeled for stable evaluation~\cite{Aslam06,Buckley04,Carterette06,Sakai08}, expert labeling remains a significant bottleneck. Another strategy, inferring implicit labels from people's behavior using the system~\cite{Joachims02}, requires large user populations. New crowdsourcing methods offer potentially large time and cost savings vs.\ traditional expert labeling, but present new quality concerns~\cite{Alonso09,Blanco11,Kazai11,Snow-emnlp08}. 

We seek methodology for estimating performance of the classifiers (relative and absolute) which is more scalable than expert labeling yet preserves high correlation with evaluation based on expert labels. While the IR community has directed considerable attention toward estimating effectiveness of ranking algorithms, relatively little work has explored estimation of classifier performance. To this end, we investigate both: 1) using only labels automatically generated by the classifiers themselves ({\em blind evaluation}); and 2) using labels obtained via crowdsourcing. In regard to blind evaluation, two broad strategies are investigated: \emph{\cas} and \emph{\sac}. \emph{\ccas} methods infer a single ``pseudo-gold'' label set by aggregating classifier labels; classifiers are then evaluated based on this single pseudo-gold label set. This strategy builds on an active area of machine learning research developing methods to aggregate multiple judgments into a single, consensus judgment set~\cite{Dawid-as79,Raykar-jmlr10}. On the other hand, \emph{\sac} methods: i) sample multiple label sets from classifier outputs, ii) evaluate classifiers on each label set, and iii) average classifier performance across label sets. To further investigate evaluation the reliability-cost tradeoff, we investigate two alternative avenues for exploiting additional labels collected via crowdsourcing: 1) direct evaluation of classifiers; or 2) supervision of \emph{combine-and-score} methods. 

To assess generality of our techniques, classifier performance is measured using four common classification metrics, with statistical significance tests establishing relative performance of the classifiers for each metric. To evaluate our techniques, we measure both score and rank correlations between estimated vs. actual classifier performance (relative and absolute). 

Figure~\ref{fig:overall} depicts our experimental framework. Experiments are conducted with ten binary classifiers submitted to the TREC 2011 Crowdsourcing Track\footnote{\url{https://sites.google.com/site/treccrowd}} investigate the following research questions: 1) Can we reliably estimate classifier performance without expert judgments? 2) How much benefit do crowd judgments provide over blind evaluation? 3) Which methods provide the best score and/or rank correlation for each labeled data condition and classifier metric of interest? 4) How robust is evaluation based on {\em\cas} methods to their labeling errors? 5) How effectively can we evaluate outlier classifiers without any judgments?

Results show high score correlation for three of the four classifier metrics considered. While crowd judgments are not seen to provide significant improvement for score correlation, they do significantly benefit rank correlation. When crowd judgments are available, we find direct evaluation on them outperforms using them to supervise {\em\cas} methods. In the blind evaluation case, simple sampling-based evaluation is typically as effective as more complicated EM, but significantly outperforms the more popular MV approach. As expected, lower quality of labels output by {\em\cas} does yield less accurate evaluation, though evaluation is reasonably tolerant of some amount of label noise. Finally, blind evaluation for outliers is surprisingly accurate, though use of crowd judgments will likely be more common in practice to achieve a more accurate ranking.

%==============================================================================
\section{Related Work} \label{sec:relatedwork}

A broad concern with human labeling, such as relevance judging, is ensuring label consistency such that systems can be effectively trained and evaluated. A decade ago, Voorhees showed that reliable evaluation of search systems could be achieved despite significant variations in the underlying relevance judgments~\cite{Voorhees00}. Soboroff et al.~\cite{Soboroff01} took this idea further: could we forgo human labeling entirely (i.e., blind evaluation) by sampling relevant documents randomly after pooling outputs from all systems participating in a shared task evaluation? While such evaluation indeed correlated positively and significantly with use of human judgments, predicted performance was less reliable for the best systems. Aslam and Savell~\cite{Aslam03} achieved comparable correlation by simply scoring each system by its mean Jaccard Coefficient over the set of retrieved documents vs.\ those retrieved by each other system. Wu and Crestani exploited similar ``reference count'' popularity, modeling expected correlation between relevance and the rank and frequency at which documents are retrieved by systems~\cite{Wu03}, similar to rank fusion~\cite{Efron09}. Several other blind evaluation methods have also been explored~\cite{Efron10,Hauff10,Shi10,Spoerri07}.

\begin{otherlanguage}{german}
Lam and Stork~\cite{Lam03} investigated correlation between label noise and classifier error. B{"u}ttcher et al.~\cite{buttcher2007reliable} investigated classifier evaluation with biased labels. Cormack et al.~\cite{cormack2009} used a pseudo-ground truth generated by spam filters to evaluate the filters, and demonstrated a lower error rate compared with labels obtained from natural sources (e.g., user labels and exhaustive adjudication by experts). In comparison to these prior works, our {\em\cas} methods approximate the pseudo-ground truth in more general fashion via aggregating classifier outputs for consensus. Cormack et al.~\cite{cormack2009} approximate the pseudo-gold labels by comparing each pair of spam filter's relative performance over some predefined measures. This approach is not directly applicable to generate consensus labels in a general way since it needs additional effort to compare each classifier's performance. Moreover, we investigate the benefit of using crowdsourced labels for the supervision of classifiers and the effect of direct evaluation of classifiers with them. 
\end{otherlanguage}

\section{Estimation Methods} \label{sec:methods}

This section describes various methods for estimating classifier performance using either no expert labels ({\em blind evaluation}) or labels via crowdsourcing. We organize methods into two general classes: \textit{\sac} and \textit{\cas}.

\textit{\csac} methods resemble the sampling approach of Soboroff et al.~\cite{Soboroff01}. Each classifier is evaluated by its average performance across some number of pseudo-gold judgment sets sampled from classifier outputs (i.e.\ we score classifiers, then combine the scores). In contrast, \textit{\cas} methods first aggregate classifier outputs into a single judgment set, then evaluate classifiers on this consensus judgment set (i.e.\ we combine the labels, then score the classifiers). The \textit{\cas} approach follows a vein of machine learning research into how to effectively integrate redundant labels produced by multiple systems~\cite{Efron09} or people~\cite{Dawid-as79} to yield consensus labels. This area has become particularly active in the context of crowdsourcing to aggregate noisy human labels~\cite{Raykar-jmlr10}. 

\emph{\csac} estimation methods are always blind (i.e.\ performed without labeled data). \emph{\ccas} methods span both unsupervised and supervised methods. For unsupervised methods, we consider Majority Vote (MV) and Expectation Maximization (EM). For supervised methods, we consider a variety of approaches: calibrated MV, Naive Bayes (NB), Support Vector Machine (SVM), Generalized Linear Model (GLM), and AdaBoost.  We also consider simply using crowdsourced labels (in lieu of expert judgments) to directly evaluate classifiers.

{\bf Notation.} $K$ input classifiers $c_{1:K}$ each label $M$ examples $x_{1:M}$ with labels $l_{mk}$. Each example $x_m$ has true label $l(x_{m})=C_m \in \{0,1\}$ for binary classification.

%-------------------------------------------------------------------------------
\subsection{Blind Methods: {\csac}} 

%..............................................................................
\subsubsection{Round-robin}

In round-robin evaluation with $K$ classifiers, we select each classifier in turn and use its labels to evaluate all $K-1$ other classifiers. Each classifier is then scored by its mean performance over all $K-1$ trials.

%..............................................................................
\subsubsection{Sampling} 

With three human assessors judging 50 topics, Voorhees~\cite{Voorhees00} performed topic-level sampling to generate new judgment sets from the space of $3^{50}$ possible combinations. In our case, for each example we randomly select a classifier to label it; with $K$ classifiers and $n$ examples, we sample our pseudo-gold from $K^n$ possible labelings. Whereas {\em round-robin} estimation is based on a sample size of $K-1$, sampling allows us arbitrarily increase the sample size $\xi$ to reduce variance (generating $\frac{\xi}{K}$ times more samples than with round-robin evaluation). We set $\xi=1000$.

%-------------------------------------------------------------------------------
\subsection{Blind Methods: {\ccas}} 

%..............................................................................
\subsubsection{Majority Vote} \label{s:unsupmv}

Majority Vote (MV) is the simplest, best-known method for generating consensus:  
simply pick the label receiving the most votes. This is equivalent to computing the average label and rounding according to a decision threshold $t$:
\begin{equation}\label{eq:mv}
\hat{l}_{MV}(m) = 1 ~\textnormal{iff.}~ \frac{1}{K} \sum_{k=1} ^{K} l_{mk} \geq t
\end{equation}
\vspace{-10pt}

where $t=\frac{1}{2}$ by default for unbiased rounding. With an even number of votes, ties are possible and broken randomly to avoid bias. More generally, the decision threshold may be varied to achieve a desired recall/specificity tradeoff (see Section~\ref{subsec:eval_measures} for definitions of classification metrics). Setting $t=0$ labels all examples as relevant (100\% recall), while $t>1$ labels all as non-relevant (100\% specificity).

%..............................................................................
\subsubsection{EM} \label{sec:em}

Expectation Maximization (EM) \cite{Dawid-as79} estimates the error rates of each classifier $c_k$ by a latent \textit{confusion matrix} [$\pi_{ij}^{(k)}$], where $ij$-th element $\pi_{ij}^{(k)}$ denotes the probability of classifier $c_k$ classifying an example to class $j$ given the true label is $i$, estimated based on each example's class membership as:

\begin{equation}\label{eq:em1}
\hat{\pi}_{ij}^{(k)}  = \frac{\sum_{m=1}^{M} T_{mi} n_{mj}^{(k)} }{ \sum_{i=1}^{C} \sum_{m=1}^{M} T_{mi} n_{mj}^{(k)} }
\end{equation} 
Indicator $n_{mj}^{(k)} = 1$ iff.\ example $x_m$ receives label $j$ from classifier $c_k$, and 
 indicator $T_{my} =1$ iff.\ $y$ is the true label for $x_m$. 
Latent class prior $p_{1:L}$ is estimated by:

\begin{equation}\label{eq:em2}
\hat{p}_i = \frac{1}{M} \sum_{m=1}^{M} T_{mi} 
\end{equation}

Since the true label for $x_m$ is unknown in the unsupervised case, EM uses a mixture of multinomials to estimate classifier accuracy. Assuming every pair of classifiers is independent, the probabilistic model likelihood can be written:
\begin{equation}\label{eq:em3}
\textit{L}(p_i, \pi_{ij}^{(k)}) = \prod_{m=1}^{M} \left( \sum_{i=1}^{C} p_i \sum_{k=1}^{K} \sum_{j=1}^{C} ({\pi}_{ij}^{(k)})^{n_{mj}^{(k)}} \right)
\end{equation}

Estimating the maximum likelihood in Equation \ref{eq:em3} is analytically intractable since it involves computing the product of a summation. However, once we get estimates for latent parameters $p_i$ and ${\pi}_{ij}^{(k)}$, we can derive new class membership $T_{mi}$ for label $l_m$ such that $T_{ml}$ = 1 if $l$ becomes the estimated true label for example $x_m$ which maximizes:

\begin{equation}\label{eq:em4}
\textit{L}(p_i, \pi_{ij}^{(k)}) = \prod_{m=1}^{M} p_i  \prod_{k=1}^{K} \prod_{j=1}^{C} ({\pi}_{ij}^{(k)})^{n_{mj}^{(k)}} .
\end{equation}

We then iteratively re-estimate latent $p_i$ and ${\pi}_{ij}^{(k)}$, and missing labels $T_{mi}$ from Equations \ref{eq:em1},  \ref{eq:em2}, and \ref{eq:em4} until convergence.

%------------------------------------------------------------------------------
\subsection{Direct Evaluation on Crowd Labels}

When crowdsourcing are available, the simplest way to use them is to evaluate classifiers on them directly. As with {\cas} approaches, the classifiers are scored against a single judgment set, though here judgments come from the crowd rather than from the classifiers.

%------------------------------------------------------------------------------
\subsection{Supervised {\ccas} Methods} \label{sec:cas}

Another way to use crowd judgments is to supervise {\cas} methods. We investigate a variety of approaches: calibrated MV, Naive Bayes (NB), Support Vector Machine (SVM), Generalized Linear Model (GLM), and AdaBoost.  

%..............................................................................
\subsubsection{Calibrated Majority Vote}

\begin{figure} 
\centering
\ifpdf
\includegraphics[width=75mm]{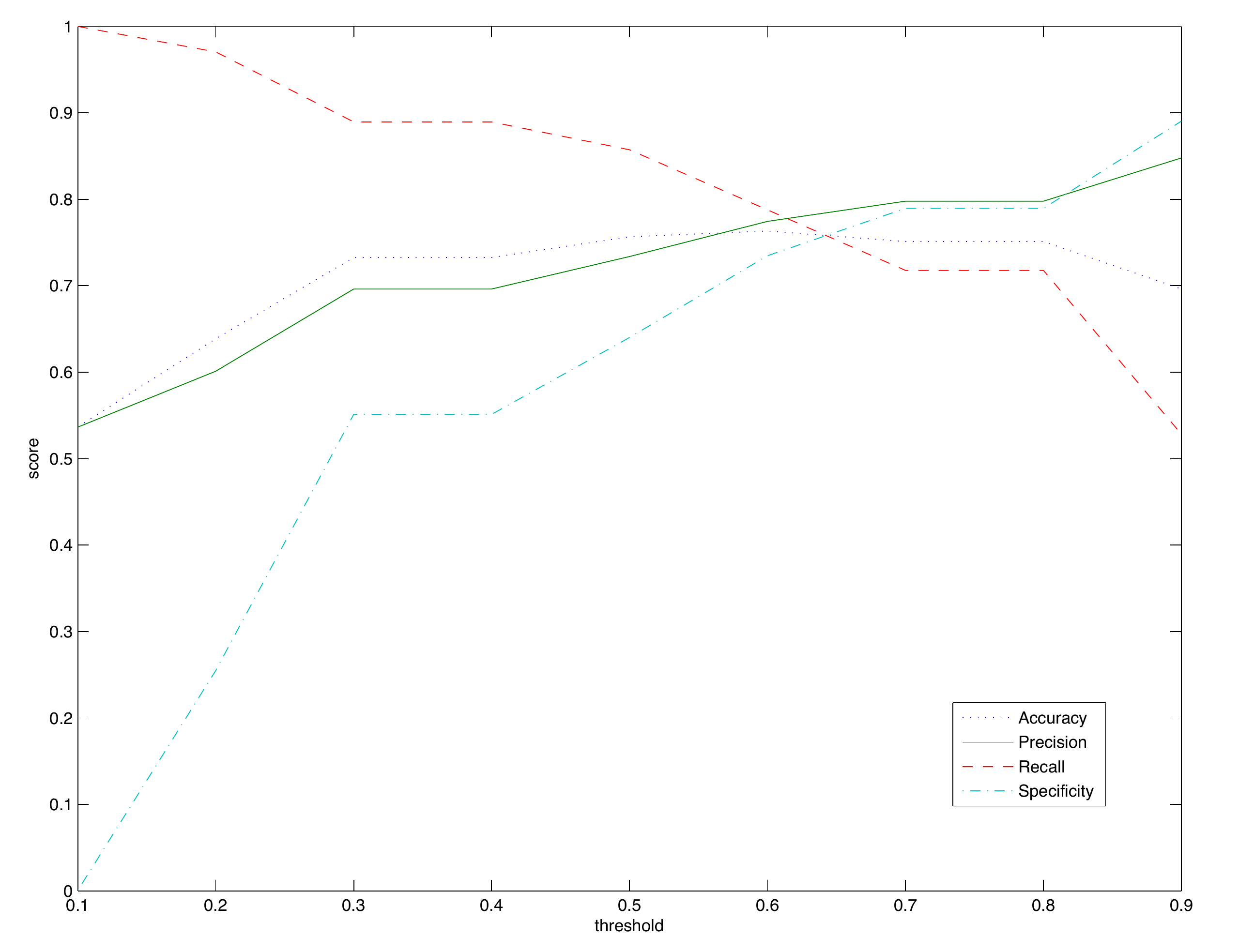}
\fi
\caption{Majority Vote (MV) aggregates input classifier labels, then follows the Bayes {\em optimal decision rule} in deciding between classes. With training data, input bias can be detected to calibrate the decision threshold $t$. Agreement between MV and expert judgments (development set) is shown for the four classification metrics from Section~\ref{subsec:eval_measures}.}
\label{fig:MV_calibration}
\end{figure}

When training labels are available (supervised setting), we can detect input bias and calibrate the decision threshold $t$ for MV (see Section~\ref{s:unsupmv} for description of unsupervised MV). Figure~\ref{fig:MV_calibration} shows the impact of varying this decision threshold for the different classifier metrics considered. While unsupervised MV uses the default $t=0.5$, in the supervised case we tune $t$ to maximize metric performance.

%..............................................................................
\subsubsection{Naive Bayes}

Naive Bayes (NB) represents another approach which exploits supervision to try to infer better pseudo-gold~\cite{Snow-emnlp08}. We assume that each input label $l_{m,1:K}$ for example $x_m$ from classifiers $c_{1:K}$ is independent. The posterior probability is calculated from the prior probability of each class $p(C=c)$ and the likelihood $P(C=c|l_{m_{1:K}})$. Likelihood is computed on the labels of each classifier. Inference is computed by: 

\begin{equation}\label{eq:nb1}
\hat{c} = \operatorname*{arg\,max}_c \left[ p(C_m=c) \prod_{k=1}^{K} P(l_{m,1:K}|C_m=c) \right]
\end{equation}

%..............................................................................
\subsubsection{Generalized Linear Model}

The generalized linear model (GLM) extends ordinary linear regression \cite{Nelder72}. For consensus label estimation, the GLM fits the given set of classifier output labels $l_{m_{1:k}}$ with training label $C_m$ for example $m$. It returns a K$\cdot$1 vector $\beta$ of coefficient estimates for a generalized linear regression of the responses in $x_m$ for classifier labels $l_{m_{1:k}}$, using the binomial distribution. We use the $logit$ link function $ln (\frac {\mu}{1-\mu})$.

%..............................................................................
\subsubsection{SVM}

A Support Vector Machine (SVM) learns a hyper-plane decision surface between classes, defined by learned ``support vectors''  which maximize the separation margin between example classes observed in training data \cite{Boser92}. We adopt a simple linear kernel. The training function identifies support vectors $s_j$, weights $\alpha_i$, and bias $b$ that are used to classify vectors $x$ according to the following equation:
\begin{equation}\label{eq:svm}
g_m(x) = \sum_{k=1}^{K} \alpha_m k(s_m, x) +b
\end{equation}
where $k$ is a kernel function, which is the dot product. If $g_m(x) \ge 0$, then $l_{svm}(m)$ is classified as relevant (1), otherwise it is classified as non-relevant (0).

%..............................................................................
\subsubsection{AdaBoost}

Boosting is a meta-algorithm which learns iteratively weak classifiers with respect to a distribution and adding them to a final strong classifier in a weighted way. Until convergence, it iteratively re-weights the data. In this work, we adopt the well-known AdaBoost algorithm~\cite{Yoav:ada}.

%==============================================================================
\section{Evaluation Methodology}

This section describes how we evaluate alternative methods for estimating classifier performance. We begin by describing the datasets used in our study. Next, we define the classifier metrics used. Following this, we describe our method for ranking classifiers given their scores and results of statistical significance tests of differences between scores. Finally, we discuss the {\em{triangle}} testing method we use to test statistical significance of differences between score and rank correlations achieved by different score estimation methods. 

%------------------------------------------------------------------------------
\subsection{Data} 
\label{subsec:dataset}

\begin{table} 
\centering
\begin{tabular}[ht]{|c |c| c| c| c|}
\hline
ID & Source & Use & {\nms}Examples{\nms} & {\nms}Labels Per Ex.\ {\nms}\\
\hline
D1 & Classifier & Input & 16,758 & 10\\ 
D2 & Crowd  & Train & 1,906& 1\\
D3 & Waterloo & Validation & 1,000 & 1\\
D4 & NIST  & Test 1& 1,000 & 1\\
\hline
\end{tabular}
\caption{Experiments are performed using four distinct sets of binary labels (D1-D4) drawn the 2011 TREC Crowdsourcing Track.} \label{table:dataset_desc}
\label{t:data}
\end{table}

Table~\ref{t:data} describes the four datasets used in our experiments (D1-D4). Data are drawn from the TREC 2011 Crowdsourcing Track, which was comprised of two distinct tasks. Task 1 involved collecting crowd judgments, while Task 2 involved aggregating crowd judgments to classify each example. $K=10$ teams participating in Task 2 classified $M=16,758$ examples, yielding the 10 classifiers to be evaluated and their outputs (D1).  

Expert judgments used come from two distinct sources. For development and tuning, we use a set of 1,865 expert labels produced by the University of Waterloo~\cite{Smucker12} (D3). For final testing, we use a balanced set of 1,000 NIST judgments (D4) on which classifiers were officially evaluated in the track. 

Crowd judgments used (D2) are drawn from Task 1 submissions. Each of 2,715 examples was labeled by $\approx$ 4-5 teams, collapsed to a single label by majority vote.

%------------------------------------------------------------------------------
\subsection{Metrics for Classifier Performance} 
\label{subsec:eval_measures}

To assess generality of methods across a range of potential metrics of interest, classifiers are evaluated on four common metrics: $tp$ is the number of true positive classifications, $fp$ false positives, $tn$ true negatives, and $fn$ false negatives:
\begin{eqnarray} 
\textnormal{Accuracy (ACC)} = \frac{tp+tn}{tp+tn+fp+fn} \\
\textnormal{Precision~(PRE)} = \frac{tp}{tp+fp} \\
\textnormal{Recall (REC)} = \frac{tp}{tp+fn} \\
\textnormal{Specificity (SPE)} = \frac{tn}{tn+fp} 
\end{eqnarray}
While many other classifier metrics could have also been studied, these four metrics represent a fair sample of potential metrics of interest for classification performed under varying operational settings. Statistical significance of observed differences is measured via a two-tailed, paired t-test.

%------------------------------------------------------------------------------
\subsection{Ranking Classifiers For Each Metric}  \label{subsec:ranking}

Because not all differences in observed metric scores are statistically significant, it would be misleading to compare rank distinctions between classifiers whose score differences are not significant. This yields (potentially conflicting) pair-wise preference constraints from which we must then induce a ranking over classifiers. While optimal ordering is NP-Hard, simple heuristics exist~\cite{Cohen99}, and we allow and model tied ranks. There is also the question of loss function: should higher-ranked preference violations be penalized more heavily or should penalties be uniform at all ranks? 

To rank classifiers based on pair-wise preferences from significance testing, we use Copeland's method in which items are ordered by the number of pairwise victories minus pairwise defeats~\cite{Copeland:pair}. Suppose three classifiers $C_{A:C}$ achieve correlation $r_{A:C}$. Assume differences of $r_A$and $r_B$, as well as $r_B$ and $r_C$, are not statistically significant, but the difference $r_A$ vs. $r_C$ is significant. The final order induced is thus $C_A > C_B > C_C$, since $C_A$ obtains one win (vs.\ $C_C$) and one tie (vs.\ $C_B$), B obtains two ties, and C obtains one tie and one loss. We adopt this method largely for its simplicity, though the impact of ordering algorithm and loss function on our evaluation will be further studied in future work.

%------------------------------------------------------------------------------
\subsection{Score and Rank Correlation Measures} \label{subsec:cor_measures}

To compare the estimation methods introduced in Section~\ref{sec:methods} for each classification metric of potential interest, we measure correlation of estimated classifier scores and ranking of classifiers vs.\ actual scores and ranks according to expert judgments. We adopt standard correlation measures: Pearson's $r$ for scores, and Spearman's $\rho$ and Kendall's $\tau$ for rankings. As a further measure of rank correlation, we measure Voorhees' ``swap'' $\%$~\cite{Voorhees00}, which estimates the probability of a discordant pair, i.e.\ the chance of any two classifiers being ranked in the wrong order. While closely related to Kendall's $\tau$, this measure ignores ties and concordant pairs. Other rank correlation measures which reflect alternative loss functions~\cite{Carterette09,Kumar10,Yilmaz08}  may be considered in future work. Note that score correlation is more sensitive than rank correlation to large errors in estimating performance on outliers (best and worst systems). For example, the worst system receives the same lowest rank whether it is 10x or 100x worse, yet its actual score may be harder to estimate.

Statistical significance of rank correlation is typically concerned with determining if a pair rankings are correlated, i.e.\ can we reject the null hypothesis $H_0$ that the two rankings are uncorrelated? For example, given 1) some estimated ranking over the classifiers and 2) the actual ranking as determined by evaluation on expert judgments, is there a significant correlation between the two rankings?  In general, we are \emph{not} interested in this question because {\em some} correlation nearly always exists. While the degree of correlation is of interest, this is what the coefficient tells us directly.   

Instead, what we want to detect is when one estimation method achieves significantly greater correlation than some other method, involving {\em triangle} significance testing between three rankings: the ranking given by evaluation on expert judgments vs.\ the other two estimated rankings. How do we test the statistical significance of differences in correlation? We are not familiar with established IR methodology for this. Letting $r_{ab}$ denoting the correlation coefficient between two sets of scores or rankings, the null hypothesis $H_0$ supposes that two observed coefficients $r_{xy}$ and $r_{xz}$ are equivalent (where $x$ denotes reference scores or rankings based on expert judgments, while $y$ and $z$ denote two sets of estimated scores or rankings). We compute the $t$ statistic for triangle significance testing following Hotelling~\cite{Hotelling40}: 
\begin{equation}\label{eq:comp_corr}
t = \frac {(r_{xy} - r_{xz}) \sqrt{(n-3)(1+r_{yz})}} { \sqrt{2(1-r_{xy}^2 - r_{xz}^2 - r_{yz}^2 + 2r_{xy}r_{xz}r_{yz})}}
\end{equation}
with n-3 degrees of freedom and $n$ being the triple ordered sample size, provided $\forall_r |r| \neq 1$. For example, with $n$ = 120,  $r_{xy}$ = 0.73,  $r_{xz}$ = 0.61, and $r_{yz}$ = 0.66, we find that $t=2.378$ with $p=0.02$, providing strong evidence for rejecting $H_0$.

%-------------------------------------------------------------------------------------------------------------------------------
\section{Results} \label{sec:results}

Recall the goal of our study is to identify effective methods for evaluating classifiers when we have no expert labels, having either no labels at all (blind evaluation) or only crowd labels. In comparing alternative methods for estimating classifier scores, we seek to identify methods whose estimated scoring and ranking of classifiers achieves high correlation with evaluation using expert labels.

We begin in Section~\ref{subsec:var_dataset} by showing variation across the label sets produced by each of the 10 classifiers. Similar to the earlier analysis of variation in assessor judgments presented by Voorhees~\cite{Voorhees00}, this analysis establishes the diversity of input label sets being used for evaluation. Next, Section~\ref{subsec:act_perfom} shows the actual scores and ranking of the 10 classifiers according to expert judgments (D3 and D4). Following this, Section~\ref{subsec:valid_result} compares the score and rank correlation achieved by the various {\em\sac} and {\em\cas} estimation methods, in supervised and unsupervised settings, on the validation set. Analysis presented here informs our selection of the best performing unsupervised and supervised methods to evaluate on the test set (D4). Next, Section~\ref{subsec:test_result} presents correlation results on the test set.

Additional analysis is presented in following sections. In Section~\ref{sec:impact}, we analyze how sensitive our evaluation of classifiers is accuracy of the labels used (in place of expert judgments). Finally, Section~\ref{sec:outliers} assesses accuracy of evaluation for outliers: the best and worst performing classifiers, whose outputs most differ from those of the other classifiers.

%------------------------------------------------------------------------------
\subsection{Agreement Between Classifiers} \label{subsec:var_dataset}

\begin{table}
\newcommand{\myeq}{\!=\!}
\centering
\begin{tabular}[ht]{ |c | c  c  c  c  c  c  c  c  c  |  }
\hline
 & C1 & C2 & C3 & C4 & C5 & C6 & C7 & C8 & C9 \\%& C10\\
\hline
C2 &.58 &&&&&&&&\\%&- & .555&.530&.691&.581&	.513&.047	&.747&	.583\\
C3 & .52 & .56 &&&&&&&\\ %& - & .448 & .504 & .510 & .474 & .025 & .565 & .520\\
C4 & .46 & .53& .45 &&&&&&\\%& - & .499 & .465 & .447 & .073 & .540 & .465 \\
C5 & .53 & .69& .50  & .50 &&&&&\\%& - &.535& .472	&.133&.695& .530\\
C6 & .54 & .58 & .51 & .47 & .54 &&&&\\%& - & .491 & .040	& .591 & .551\\
C7 & .49 & .51 & .47 & .45 & .47 & .49 &&&\\%& - & .039 & .521 & .498\\
C8 & .02 & .05 & .03 & .07 & .13 & .04 & .04 &&\\%& -  & .081 & .021 \\
C9 & .59 & .75 & .57 & .54 & .70 & .59 & .52 & .08 &\\%& - & .591\\
C10 & .56 & .58& .52 & .47 & .53 & .55 & .50 & .02 & .59 \\%& - \\
\hline
\end{tabular}
\caption{Jaccard coefficient between relevant documents sets identified by each D1 classifier (called ``overlap'' by Voorhees~\cite{Voorhees00} and \emph{SysSimilarity} by Aslam et al.~\cite{Aslam03} ($\bar{x}\myeq.44$, $s\myeq.21$, $max\myeq.75$, $min\myeq.02$).} 
\label{table:Jaccard}
\end{table}

We begin by measuring overlap across the sets of relevant documents identified by the 10 different classifiers considered, providing a measure of their diversity. We note that Voorhees' ``overlap''~\cite{Voorhees00} is the \emph{Jaccard} similarity coefficient computed between label sets (the size of the intersection over the size of their union). Pair-wise results are shown in Table~\ref{table:Jaccard}. 
Overall, we observe overlap with mean $\bar{x}=0.442$ and standard deviation $s=0.205$ across the classifier set, relatively similar to overlap levels seen in earlier studies with human assessors. Crucially, though, we observe that classifier $C8$ exhibits extremely low overlap vs.\ all other classifiers. At this point in discussion, we are not yet certain whether $C8$ is far better or far worse than the rest of the pack, though intuition suggests the latter is more likely. 

In addition to looking at overlap between classifiers, we also inspected Fleiss $\kappa$, a widely used measure for annotator agreement between a fixed number of raters \cite{Fleiss:kappa}. We observe $\kappa=0.27$ between the 10 classifiers, representing low but ``fair'' (not chance) agreement. If we exclude the outlier $C8$, however, we observe far higher $\kappa=0.50$ over the remaining 9 classifiers, indicating ``moderate'' agreement. 

%------------------------------------------------------------------------------
\subsection{Actual Classifier Performance} \label{subsec:act_perfom}

\begin{table}[t]
\newcommand{\myr}{\!\!\!Rank\!}
\centering
\scalebox{0.95}{%
\begin{tabular}[h]{|l|ll|ll|ll|ll| }
\hline

	&ACC	&\myr	&PRE	&\myr	&REC	&\myr	&SPE	&\myr\\ \hline
C1	&0.74	&1*	&0.61	&2*	&0.76	&3*	&0.72	&2*\\
C2	&0.68	&*6	&0.55	&*6	&0.80	&2	&0.62	&7\\
C3	&0.73	&1*	&0.61	&2*	&0.71	&6	&0.73	&2*\\
C4	&0.47	&9	&0.37	&9	&0.64	&*7	&0.38	&9\\
C5	&0.67	&*6	&0.54	&*6	&0.62	&*7	&0.69	&6\\
C6	&0.72	&1*	&0.59	&2*	&0.75	&3*	&0.70	&2*\\
C7	&0.74	&1*	&0.65	&1	&0.65	&*7	&0.79	&1\\
C8	&0.29	&10	&0.16	&10	&0.21	&10	&0.33	&10\\
C9	&0.67	&*6	&0.53	&*6	&0.85	&1	&0.57	&8\\
C10	&0.73	&1*	&0.61	&2*	&0.77	&3*	&0.71	&2*\\ \hline
$\bar{x}$&0.64	&	&0.52	&	&0.67	&	&0.63	&\\ 
$s$	&0.15	&	&0.15	&	&0.18	&	&0.16	&\\ \hline
\end{tabular}}
\caption{Actual performance of classifiers on validation set (D3) expert judgments for all four metrics considered (mean $\bar{x}$ and standard deviation $s$ are also shown). Tied ranks according to statistical significance testing are indicated by * (95\% confidence).} 
\label{table:valid_gold_eval}
\end{table}

\begin{table}[t]
\newcommand{\myr}{\!\!\!Rank\!}
\centering
\scalebox{0.95}{%
\begin{tabular}[ht]{|l|ll|ll|ll|ll| }
\hline
	&ACC	&\myr	&PRE	&\myr	&REC	&\myr	&SPE	&\myr\\  \hline
C1	&0.69	&1*	&0.65	&2*	&0.79	&*3	&0.58	&2*\\
C2	&0.61	&*7	&0.57	&*7	&0.88	&2	&0.33	&8\\
C3	&0.68	&1*	&0.65	&2*	&0.76	&7*	&0.59	&2*\\
C4	&0.69	&1*	&0.66	&2*	&0.79	&*3	&0.60	&2*\\
C5	&0.53	&9	&0.52	&9	&0.82	&*3	&0.23	&10\\
C6	&0.66	&6	&0.64	&2*	&0.76	&7*	&0.56	&7\\
C7	&0.70	&1*	&0.68	&1	&0.75	&7*	&0.64	&1\\
C8	&0.37	&10	&0.25	&10	&0.14	&10	&0.60	&2*\\
C9	&0.60	&*7	&0.57	&*7	&0.91	&1	&0.30	&9\\
C10	&0.69	&1*	&0.66	&2*	&0.80	&*3	&0.59	&2*\\ \hline
$\bar{x}$	&0.62	&	&0.58	&	&0.74	&	&0.50	&\\ 
$s$	&0.11	&	&0.13	&	&0.22	&	&0.15	&\\ \hline
\end{tabular}}
\caption{Actual performance of classifiers on test set (D4) expert judgments for all four metrics considered (mean $\bar{x}$ and standard deviation $s$ are also shown). Tied ranks according to statistical significance testing are indicated by * (95\% confidence).} 
\label{table:NIST_gold_rawdata}
\end{table}

To measure our accuracy of estimating classifier performance, we must first compute the actual performance of classifiers according to expert judgments. We begin with the validation set (D3) and measure actual performance of the 10 classifiers. Results appear in Table~\ref{table:valid_gold_eval}. We see results on all four classification metrics, as well as sample mean $\bar{x}$ and standard deviation $s$. Tied ranks according to two-tailed paired-t test are indicated by * (use of *M vs.\ N* merely groups equivalent rank values vertically). Classifiers perform rather comparably, though classifier C8 performs far worse than other classifiers for all metrics but specificity. 

Table~\ref{table:NIST_gold_rawdata} shows the actual performance achieved by all classifiers on the test set (D4) based on expert judgments. 

%------------------------------------------------------------------------------
\subsection{Correlation Results on Validation Set (D3)} \label{subsec:valid_result}

This section presents our validation set (D3) results, centered on the score and rank correlation achieved by each estimation method introduced in Section~\ref{sec:methods}. The statistical significance of differences in correlation achieved by alternative methods is measured, with significance reported at 95\% confidence or higher (see Section~\ref{subsec:cor_measures} for details). Analysis of validation set results provides our initial insights into research questions, including identification of best performing methods to be evaluated on the test set (D4, Section~\ref{subsec:test_result}). 

\begin{figure}
\centering
\ifpdf
\includegraphics[width=85mm, height=65mm]{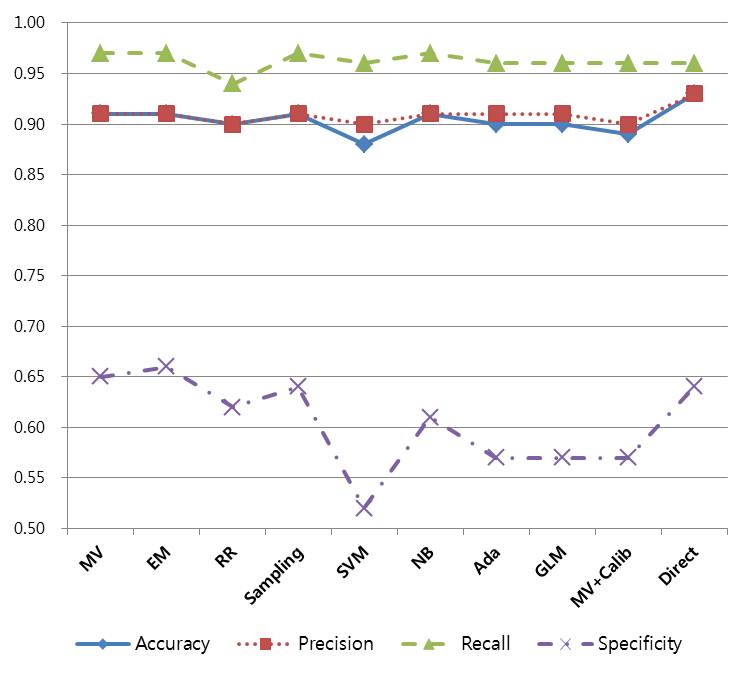}
\fi
\caption{Validation set (D3) score correlation (Pearson's $r$) achieved by estimation methods from Section~\ref{sec:methods}. The x-axis is labeled by method; the y-axis shows the correlation achieved. Results for each metric from Section~\ref{subsec:eval_measures} are connected by a different line. While correlation is high across methods for ACC, PRE, and REC metrics ($r > 0.9$ typically), SPE correlation is much lower ($0.5 < r < 0.67$).} 
\label{fig:valid_pearson}
\end{figure}

\begin{figure*}[t]
\begin{center}$
\begin{array}{c}
\subfigure[Spearman's $\rho$ rank correlation on the validation set (D3)]{
\ifpdf %ML
\includegraphics[width=80mm, height=60mm]{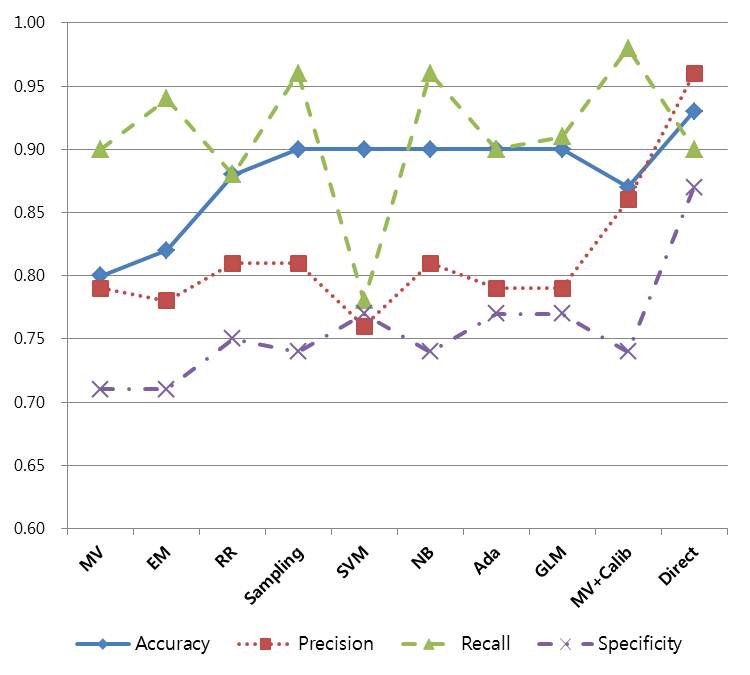}
\fi
\label{fig:subfig1}
}\\
\subfigure[Kendall's $\tau$ rank correlation on the validation set (D3)]{
\ifpdf %ML
\includegraphics[width=80mm, height=60mm]{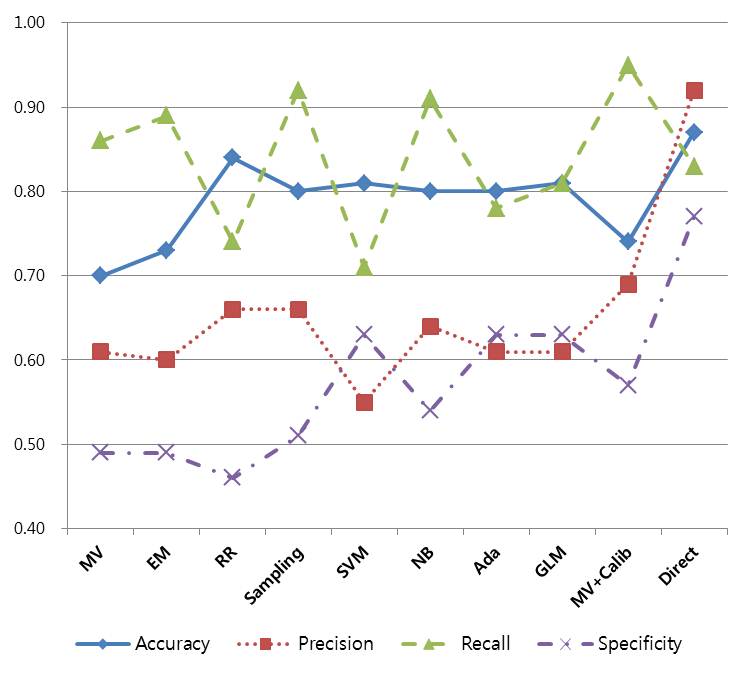}
\fi
\label{fig:subfig2}
}\\
\end{array}$
\caption[Optional caption for list of figures]{Validation set (D3) rank correlation, Spearman's $\rho$ \subref{fig:subfig1} and Kendall's $\tau$ \subref{fig:subfig2}, achieved by Section~\ref{sec:methods} methods. The x-axis is labeled by method. From left-to-right: the four blind methods: MV and EM (C\&S) followed by RR and Sampling (S\&C); next the five supervised C\&S methods: SVM, NB, Ada, GLM, and Calibrated MV; finally crowd-based direct evaluation. The y-axis shows the correlation achieved by each method. Results for each classifier metric, ACC, PRE, REC, and SPE, are plotted by a different line. } 
\label{fig:valid_rank}
\end{center}
\end{figure*}

{\bf Summary of validation set findings.} 1) Can we reliably estimate classifier performance? 2) Do crowd judgments enable us to do this significantly more effectively? We observe high Pearson score correlation of 0.9 or greater over the 10 classifiers for three of the four classifier metrics considered. Crowd judgments are not seen to provide significant improvement. For rank correlation, however, we observe Spearman correlation between 0.87 and 0.96 with crowd judgments, but a lower 0.75 to 0.95 without judgments. 3) Which methods provide the best score and/or rank correlation for each labeled data condition and classifier metric of interest? Figures~\ref{fig:valid_pearson} and \ref{fig:valid_rank} answer this question via method vs. metric plots of raw correlation levels, while Table~\ref{table:valid_cor_rank} ranks the alternative methods based on statistical significance testing of differences in correlation achieved.

% pearson p on validationset results

% spearman's rho and kendall's tau on validation set

{\bf Details.} Figure~\ref{fig:valid_pearson} shows Pearson's $r$ score correlation achieved by each method from Section~\ref{sec:methods} for all four metrics: ACC, PRE, REC, and SPE. Correlation is quite high across methods and metrics except for specificity, which is much lower. No significant benefit is seen from use of crowd judgments vs.\ blind evaluation, with comparable correlation achieved by the best performing methods in each class (direct evaluation and EM, respectively). 

Figure~\ref{fig:valid_rank} shows rank correlation (Spearman's $\rho$ \subref{fig:subfig1} and Kendall's $\tau$ \subref{fig:subfig2}) achieved by the different methods from Section~\ref{sec:methods} on the validation set (D3). The plot for Swap \% rank correlation was quite similar to Kendall's $\tau$, so we omit it to simplify our presentation. Unlike score correlation, with rank correlation we do see significant improvement from use of crowd judgments. For example, we observe Spearman correlation between 0.87 and 0.96 with crowd judgments. Without judgments, correlation achieved by blind methods spans a wider range, from around 0.75 to 0.95. With Kendall's $\tau$, we see around 0.78-0.92 correlation across metrics for direct evaluation (with crowd judgments), while blind EM ranges from 0.5-0.9 correlation across metrics.

% ranks of validation-set based results
\begin{table*}[h]
%\begin{sidewaystable}
\newcommand{\z}{$\dagger$}
\newcommand{\metrics}{&\!\!ACC\!\!&\!PRE\!\!&\!REC\!\!&\!SPE\!\!}
\centering
\scalebox{0.66}{%
\begin{tabular}{|c| c| c|cccc||  c c c c | c c c c | cccc|}
\hline
	&	&	&\multicolumn{4}{|c||}{\bf Score correlation} & \multicolumn{12}{|c|}{\bf Rank correlation} \\ \cline{4-19}
& & & \multicolumn{4}{|c||}{\bf Pearson $r$} & \multicolumn{4}{|c}{\bf Spearman $\rho$}&  \multicolumn{4}{|c}{\bf Kendall $\tau$} &  \multicolumn{4}{|c|}{\bf Swap \%} \\
\hline
\!\!Blind\!\! &\!\!Type\!\! 	&\!\!Method\!\!	\metrics	\metrics	\metrics	\metrics	\\\hline
Yes	&C\&S	&MV			&2	&1	&1	&1	&10	&6	&5	&9	&10`	&6	&5	&7	&10	&6	&6	&7\\
	&C\&S	&EM			&2	&1	&1	&1	&9	&6	&4	&9	&8	&6	&4	&7	&8	&6	&3	&7\\\cline{2-19}
	&S\&C	&RR			&2	&1	&1	&5	&2	&3	&9	&2	&1	&3	&9	&7	&1	&3	&9	&10\\
	&S\&C	&Sampling		&2	&1	&1	&1	&2	&3	&2	&2	&3	&3	&2	&7	&3	&3	&3	&7\\\hline
No	&C\&S	&SVM			&2	&1	&1	&10	&2	&10	&10	&2	&3	&10	&10	&2	&3	&10	&9	&2\\
	&C\&S	&NB			&2	&1	&1	&5	&2	&3	&2	&2	&3	&3	&3	&6	&3	&3	&3	&6\\
	&C\&S	&Ada			&2	&1	&1	&5	&2	&6	&5	&2	&3	&6	&8	&2	&3	&6	&8	&2\\
	&C\&S	&GLM			&2	&1	&1	&5	&2	&6	&5	&2	&3	&6	&6	&2	&3	&6	&7	&2\\
	&C\&S	&MV+Calib(0.6)	&2	&1	&1	&5	&8	&2	&1	&2	&8	&2	&1	&5	&8	&2	&1	&5\\\cline{2-19}
	&-	&Direct-eval		&1	&1	&1	&1	&1	&1	&5	&1	&1	&1	&6	&1	&1	&1	&1	&1\\
\hline
\end{tabular}}
\caption{Validation set (D3) results show relative correlation achieved by alternative estimation methods introduced in Section~\ref{sec:methods}. Correlation is measured between predicted scores and derivative ranking of classifiers, vs. actual scores and derivative ranking based on expert judgments. Note that ranks shown refer to estimation methods, not the classifiers, indicating the relative correlated achieved by alternative methods. A higher ranked method achieves statistically significantly better correlation, with tied ranks indicating equivalent correlation. Ranks reflect triangle significance testing of correlation at 95\% confidence (see Section~\ref{subsec:cor_measures}).}
\label{table:valid_cor_rank}

%\end{sidewaystable}
\end{table*}

Table~\ref{table:valid_cor_rank} ranks methods (with ties) for all four classifier metrics according degree of correlation achieved with evaluation on expert judgments in the validation set (D3). In comparison to raw correlation values shown in Figures~\ref{fig:valid_pearson} and \ref{fig:valid_rank}, method rankings in Table~\ref{table:valid_cor_rank} reflect statistical significance testing, i.e. a higher ranked method achieves statistically significantly better correlation, with tied ranks indicating equivalent correlation. The best method using crowd judgments, direct evaluation, is shown in the bottom row. Without judgments, sampling achieves lower correlation than direct evaluation, particularly for the SPE metric, as noted earlier. Based on these results, we also select two {\cas} methods to evaluate in final testing: EM as a blind method and NB as a supervised method.

% top methods on test set 

%------------------------------------------------------------------------------
\subsection{Correlation Results on Test Set (D4)} \label{subsec:test_result}

%\begin{sidewaystable}
\begin{table*}[hp]
\newcommand{\z}{$\dagger$}
\newcommand{\metrics}{&\!\!ACC\!\!&\!PRE\!\!&\!REC\!\!&\!SPE\!\!}
\centering
\scalebox{0.64}{%
\begin{tabular}{|c| c| c|cccc||  c c c c | c c c c | cccc|}
\hline
& & &\multicolumn{4}{|c||}{\bf Score correlation} & \multicolumn{12}{|c|}{\bf Rank correlation} \\ \cline{4-19}
& & & \multicolumn{4}{|c||}{\bf Pearson $r$} & \multicolumn{4}{|c}{\bf Spearman $\rho$}&  \multicolumn{4}{|c}{\bf Kendall $\tau$} &  \multicolumn{4}{|c|}{\bf Swap \%} \\
\hline
\!\!Blind\!\! &\!\!Type\!\! 	&\!\!Method\!\!	\metrics	\metrics	\metrics	\metrics	\\\hline
Yes	&C\&S&EM		&.91	& .98&.99&.54		&.69	&.80		&{\bf.83}	&.29	&.60&.68&{\bf.72}&.31	&15.6&11.1&8.9&24.4\\
	&S\&C&Sampling	&.92	& .98&.99&.59*	&{\bf.79}\z& .84*	& {\bf.80}	&.32	&.66\z	&.69	&{\bf.71}	&.33	&13.3	&11.1	&11.1	&24.4 \\\hline
No	&C\&S&NB		&.{\bf 94}*	& .98	&.98	&{\bf .66}\z	& {\bf.80}\z	&.84*	&.80	&.34	&.69\z	&.72	&{\bf.69}	&.35	&{\bf11.1}*	&8.9	&11.1	&22.2\\
	&-&Direct		&.90	&.96	&.98	&.67\z	&{\bf.78}\z	&{\bf.90}\z	&.77\z	&.46\z	&.68\z&{\bf.86}\z	&{\bf.70}	&.43\z	&13.3	&{\bf2.2}\z	&{\bf6.7}&20.0*\\\hline
\end{tabular}}
\caption{{\bf Test set results} (D4): correlation of alternative prediction methods vs.\ actual scores and ranks from NIST judgments. 
Blind methods use no labeled data while non-blind methods make use of crowd labels (D2). C\&S denotes a ``{\ccas}'' approach while S\&C denotes a ``{\csac}'' approach. {\em Direct} denotes evaluating classifiers directly on crowd labels (D2). Statistical significance of correlation differences between EM and other prediction methods is indicated by * (95\% confidence) and \z~(99\% confidence).} 
 \label{table:NIST_corr}
\end{table*}
%\end{sidewaystable}

Informed by validation set results (Section~\ref{subsec:valid_result}), we selected a top performing method from each method group considered and evaluated these methods on NIST gold judgments (D4). For blind evaluation with {\em\cas}, EM is reported. For supervised {\em\cas}, we report NB. Sampling is reported for {\em\sac}. In the crowd label (D2) condition, we again compare use of these labels for supervised {\em\cas} vs.\ simple direct evaluation.

As shown in Table~\ref {table:NIST_corr}, most methods show strong score correlation with evaluation based on NIST judgments. For rank correlation, all of sampling, NB, and direct evaluation methods achieve significantly higher correlation than EM. In the absence of expert judgments, {\em\sac} methods show better correlations rather than unsupervised {\em\cas} methods, consistent with earlier results on the validation set. With crowd judgments, direct evaluation achieved significantly higher correlation than NB on both PRE and SPE for Kendall $\tau$ rank correlation.

%------------------------------------------------------------------------------
\subsection{Quality of Pseudo-Gold vs. Correlation} \label{sec:impact}

The \emph{\cas} approach offers a potentially valuable separation of concerns: let machine learning research tackle the label aggregation problem~\cite{Raykar-jmlr10}, then use the output ``pseudo-gold'' as if it were expert gold. Intuitively, the more accurate the pseudo-gold, the better this strategy can be expected to work. A key concern, however, is how robust evaluation on pseudo-gold will be to impurity (e.g. labeling errors)? To investigate this, we analyze the relationship between: a) labeling quality of alternative {\cas} methods, vs.\ b) derivative correlation between predicted vs. actual scores and ranks of classifiers. We quantify the relationship between label quality and correlation by computing another, {\em secondary} Pearson $r$ correlation between label quality and each of the {\em primary} correlations in evaluation accuracy achieved by the estimation method (the four different correlation measures discussed thus far).

For each classification metric (ACC, PRE, REC, SPE), we compute the secondary Pearson $r$ correlation between the label quality vs.\ primary correlation over the 7 different {\em \cas} methods considered. Table~\ref{table:pseudo_corr} shows the results on the test set (D4). Table~\ref{table:pseudo_corr}'s caption further details our process to generate these results. Note that Swap \% values in Table~\ref{table:pseudo_corr} are negative since it is a measure of error while performance metrics measure quality, hence they are inversely correlated, as the negative values indicate. 

\begin{table}[h]
\centering
\begin{tabular}{| c| c| c| c| c |  }
\hline
Method &ACC	&PRE	&REC	&SPE	\\  \hline
MV	&0.691	&0.642	&0.864	&0.518\\
EM	&0.692	&0.671	&0.752	&0.632\\
\hline
MV+Calib(0.6)	&0.692	&0.671	&0.752	&0.632\\
Ada	&0.691	&0.666	&0.708	&0.614\\
NB	&0.692	&0.661	&0.790	&0.594\\
GLM	&0.690	&0.663	&0.774	&0.606\\
SVM	&0.676	&0.651	&0.762	&0.590\\
\hline
\end{tabular}
\caption{Test set (D4) quality of pseudo-gold labels output by unsupervised and supervised {\em\cas} methods (Rows 1-2 vs.\ Rows 3-7, respectively.} \label{table:NIST_gold_eval}
\end{table}

\begin{figure}
\centering
\ifpdf
\includegraphics[width=95mm, height=75mm]{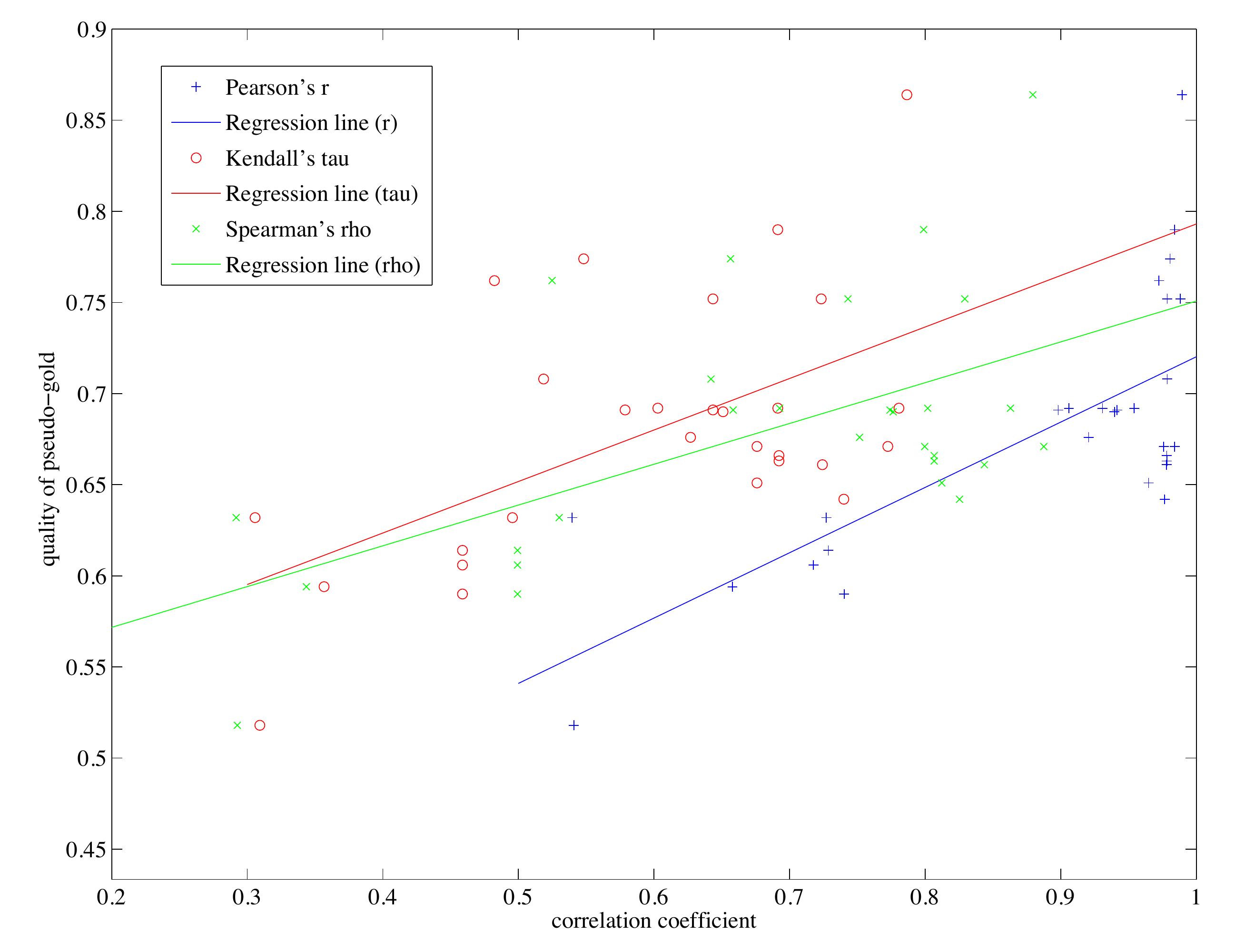}
\fi
\vspace{-10pt}
\caption{How the quality of pseudo-gold labels produced by \emph{\cas} methods impacts effectiveness in predicting classifier scores and rankings. Strength of correlation is measured by Pearson $r$between a quality of pseudo-gold and a correlation coefficient achieved by the pseudo-gold.} 
\label{fig:pseudo_corr}
\end{figure}

Results shown in Figure~\ref{fig:pseudo_corr} empirically validate our intuition: the performance of the \emph{\cas} method (i.e.\ the quality of pseudo-gold it produces) is strongly correlated with how effectively we can predict classifier scores and correctly rank classifiers. Y-axis in the figure means quality of each pseduo-ground truth. X-axis in the figure indicates a correlation coefficient of predicting classification performance achieved by the pseudo-gold. The regression line shows correlation strength, which is 0.71 (blue) for Pearson's $r$, 0.56 for Kendall's $\tau$, and 0.57 for Spearman's $\rho$. As our expectation, the higher quality of pseudo-gold achieves the higher correlation coefficient, which demonstrates that there is strong correlation between pseudo-gold quality and its accuracy of classifier performance prediction. Consequently, we assure that it is critical to generate a highly accurate pseudo-gold labels for accurately evaluating classifier performance in the absence of ground truth.

As expected, the general trend shows label quality is strongly correlated with how effectively we can predict classifier scores and derivative ranks. The ACC and PRE differences between 7 pseudo-gold labels are too tiny to show strong correlations. However, the differences of REC and SPE between the given pseudo gold labels are reasonably large, thus there are stronger correlations than in the former two metrics.

\begin{table}[t]
\centering
\begin{tabular}{|l|llll|  }
\hline
Metric & Pearson $r$ & Spearman $\rho$ & Kendall $\tau$ & Swap \% \\
\hline
ACC & 0.34	&0.38	&0.22	&-0.24 \\
PRE & 0.32	&0.28	&0.32	&-0.30	\\
REC & 0.71	&0.57	&0.64	&-0.65 \\
SPE & 0.62	&0.66	&0.69	&-0.49\\
\hline
\end{tabular}
\caption{{\em\ccas} methods generate pseudo-gold for evaluating classifiers. How robust is evaluation to errors in the pseudo-gold? Table~\ref{table:NIST_gold_eval} shows the quality of pseudo-gold labels output by the 7 methods for each metric. 
 Evaluating classifiers on pseudo-gold vs.\ expert judgments yields 4 correlation values per method for each metric. Re-grouping these values yields 7 correlation values per correlation measure for each metric. Finally, we compute Pearson $r$ correlation between the 7 label quality scores vs.\ the 7 correlation values for each metric.  Results are shown in this Table for the test set (D4). Note that Swap \% is an error measure, so is inversely related to label quality (smaller $r$ values are better). 
} 
\vspace{-10pt}
\label{table:pseudo_corr}
\end{table}

%------------------------------------------------------------------------------
\subsection{Detecting Outliers in Blind Evaluation} \label{sec:outliers}

A predominant concern in prior blind evaluation studies has been difficulty predicting performance of the best systems, since the uniqueness which distinguishes them is not ``affirmed'' by other systems~\cite{Aslam03,Soboroff01}. Moreover, success of blind evaluation is likely limited by overall quality of the system pool: if all of the systems are terrible, we cannot expect to derive much value from their outputs.

\begin{table}[ht]
\centering
\begin{tabular}{|l| l l l l |  }
\hline
	&ACC*	&PRE	&REC	&SPE\\ \hline
\hline
Actual Rank & \multicolumn{4}{|c|}{Predicted Rank}\\
\hline
1 (best)	&1	&4	&1	&4\\
2	&1	&1	&2	&10\\
3	&6	&1	&3	&3\\ \hline
10 (worst)  & 10 & 10  & 10 &  9\\
\hline
\end{tabular}
\caption{Rows 1-3: EM predicted ranks for the top-3 classifiers per metric according to NIST judgments (D4). Note that the top-3 classifiers for ACC actually tied for rank 1. Row 4 shows EM predicted rank for the worst classifier based on NIST judgments.}
\label{table:top3_corr}
\vspace{-10pt}
\end{table}

This section focuses exclusively on blind evaluation, analyzing how well we are able to predict performance of the best and worst classifiers when no judgments are available. We begin by analyzing our ability to accurately rank the top-3 classifiers according to NIST judgments (test set D4). At the other extreme, we analyze effectiveness of identifying the worst performer, classifier {\em C8}. Refer to Table~\ref{table:NIST_gold_rawdata} for actual performance achieved by all classifiers on the test set. As a representative estimation method for blind evaluation, we arbitrarily select EM (Section~\ref{sec:em}) and compare its predicted scores and ranks to the actual scores and ranks.

Table~\ref{table:top3_corr} presents our results. Rows 1-3 show the predicted ranks for the top 3 classifiers (according to actual evaluation on NIST judgments). For example, we see the best classifier for PRE is only predicted at rank 5 by EM, while the 2nd-ranked classifier is predicted at rank 1. Note that for ACC, the top-3 classifiers are actually tied (equivalent). 

Can the best classifiers be accurately predicted with blind evaluation?  Results here indicate the answer depends on the classification metric of interest. For REC, the EM predicted ranking perfectly matches the actual ranking. With ACC, the top-2 systems are correctly predicted but the 3rd system (tied with the others) is mistakenly predicted at rank 6. For PRE, the 2nd and 3rd best systems are nearly predicted, while the best system is predicted at rank 5 (noted earlier). SPE is by far seen to be the most difficult metric to predict accurately, consisitent with our earlier findings for all systems on the validation set (Figures~\ref{fig:valid_pearson} and \ref{fig:valid_rank}). 

Can the worst classifier be accurately detected with blind evaluation?  Row 4 in Table~\ref{table:top3_corr} shows predicted vs. actual rank of the worst performing classifier for each metric ({\em C8} for all metrics but SPE). For ACC, PRE, and REC, the predicted rank matches the actual rank, while for SPE the predicted rank was off by one. Therefore, our ability to identify the worst system is fairly consistent with our ability to identify the best systems in depending on the metric of interest.

%==============================================================================

\section{Conclusion} \label{sec:conclusion}

This paper reported our investigation into methods for evaluating classifiers using either no judgments ({\em blind evaluation}) or crowd judgments. We pursued two general strategies. \emph{\ccas} methods aggregated classifier outputs into a single pseudo-gold judgment set on which classifiers were scored. \emph{\csac} methods scored classifiers on multiple sets of judgments sampled from classifier outputs, then averaged performance across judgment sets. In the case of crowd judgments being available, we explored two approaches for exploiting them: either direct evaluation of classifiers or supervising \emph{combine-and-score} methods. Classifiers were scored on four standard metrics and then ranked based on statistically significant differences. Experiments were conducted with 10 classifiers developed independently by teams participating in the TREC 2011 Crowdsourcing Track. 

To evaluate our methods, we reported score and rank correlation measures comparing actual classifier performance vs.\ predicted performance. To correctly interpret differences in correlation observed, we utilized Hotelling's triangle testing approach~\cite{Hotelling40} (Section~\ref{subsec:cor_measures}) which was originally proposed by Hotelling and investigated by Steiger in~\cite{Steiger:corr}; we were not familiar with established methodology in the IR community for such testing. 

In regard to research questions (Section~\ref{sec:introduction}), results showed high score correlation for three of the four classifier metrics considered. While crowd judgments were not seen to provide significant improvement for score correlation, they did significantly benefit rank correlation. When crowd judgments were available, we found that direct evaluation on them outperformed their use to supervise {\em\cas} methods. In the blind evaluation case, simple round-robin evaluation was typically as effective as more complicated EM, but significantly outperformed the more popular MV approach. As expected, lower quality of labels output by {\em\cas} did yield less accurate evaluation, though evaluation was reasonably tolerant of some amount of label noise, particularly in regard to ranking classifiers based on accuracy. Finally, blind evaluation for outliers was surprisingly accurate, though imperfections still remain; there is no silver bullet here, but a fairly effective approximation of evaluation based on expert judgments. Crowd judgments can further improve this approximation at some additional cost, but likely still cheaper than collecting expert judgments.

Future work will perform comparative studies across other datasets to better assess robustness of findings, consider alternative methods for inducing a ranking from systems scores (and significance tests), as well as impact of other rank correlation measures~\cite{Carterette09,Kumar10,Yilmaz08}.

%==============================================================================
\bibliographystyle{plain}
\bibliography{ml12_jung}

%%%%%%%%%%%%%%%%%%%%%%%%%%%%%%%%%%%%%%%%%%%%%%%%%%%%%%%%%%%%%%%%%%%%%%%%%%%%%%%
\end{document}